\PassOptionsToPackage{table}{xcolor}
\documentclass[sigconf, screen]{acmart}
\AtBeginDocument{%
  }

\setcopyright{acmlicensed}
\copyrightyear{2025}
\acmYear{2025}
\acmDOI{XXXXXXX.XXXXXXX}
\acmConference[Conference acronym 'XX]{Make sure to enter the correct
  conference title from your rights confirmation email}{June 03--05,
  2018}{Woodstock, NY}
\renewcommand\footnotetextcopyrightpermission[1]{}
\acmISBN{978-1-4503-XXXX-X/2018/06}
\usepackage{multirow}
\usepackage{tabularx}
\definecolor{mygray}{gray}{0.9}  % 0 = 黑，1 = 白，0.9 是浅灰
\usepackage{pifont}
\usepackage{bm}
\usepackage{overpic}
\usepackage{graphicx} % 需在导言区引入
\usepackage{adjustbox} % 如需对齐控制

\begin{document}

\title{GRACE: Estimating Geometry-level 3D Human-Scene Contact from 2D Images}

\author{Chengfeng Wang}
\affiliation{%
  \institution{University of Science and Technology of China}
  \city{Hefei}
  \country{China}
}
\email{WCF@mail.ustc.edu.cn}

\author{Wei Zhai}
\affiliation{%
  \institution{University of Science and Technology of China}
  \city{Hefei}
  \country{China}
}
\email{wzhai056@ustc.edu.cn}

\author{Yuhang Yang}
\affiliation{%
  \institution{University of Science and Technology of China}
  \city{Hefei}
  \country{China}
}
\email{yyuhang@mail.ustc.edu.cn}

\author{Yang Cao}
\affiliation{%
  \institution{University of Science and Technology of China}
  \city{Hefei}
  \country{China}
}
\email{forrest@ustc.edu.cn}

\author{Zhengjun Zha}
\affiliation{%
  \institution{University of Science and Technology of China}
  \city{Hefei}
  \country{China}
}
\email{zhazj@ustc.edu.cn}

\begin{abstract}
Estimating the geometry level of human-scene contact aims to ground specific contact surface points at 3D human geometries, which provides a spatial prior and bridges the interaction between human and scene, supporting applications such as human behavior analysis, embodied AI, and AR/VR. To complete the task, existing approaches predominantly rely on parametric human models (e.g., SMPL), which establish correspondences between images and contact regions through fixed SMPL vertex sequences. This actually completes the mapping from image features to an ordered sequence. However, this approach lacks consideration of geometry, limiting its generalizability in distinct human geometries. In this paper, we introduce GRACE (\textbf{G}eometry-level \textbf{R}easoning for 3D Hum\textbf{A}n-scene \textbf{C}ontact \textbf{E}stimation), a new paradigm for 3D human contact estimation. GRACE incorporates a point cloud encoder–decoder architecture along with a hierarchical feature extraction and fusion module, enabling the effective integration of 3D human geometric structures with 2D interaction semantics derived from images. Guided by visual cues, GRACE establishes an implicit mapping from geometric features to the vertex space of the 3D human mesh, thereby achieving accurate modeling of contact regions. This design ensures high prediction accuracy and endows the framework with strong generalization capability across diverse human geometries. Extensive experiments on multiple benchmark datasets demonstrate that GRACE achieves state-of-the-art performance in contact estimation, with additional results further validating its robust generalization to unstructured human point clouds.
\end{abstract}

\ccsdesc[500]{Computing methodologies~Scene understanding}
\ccsdesc[300]{Human-centered computing}
\keywords{Human-Scene Contact Estimation; 3D Computer Vision; Multimodal Learning}


\maketitle

\begin{figure}[t]
	\centering
	\includegraphics[width=1\linewidth]{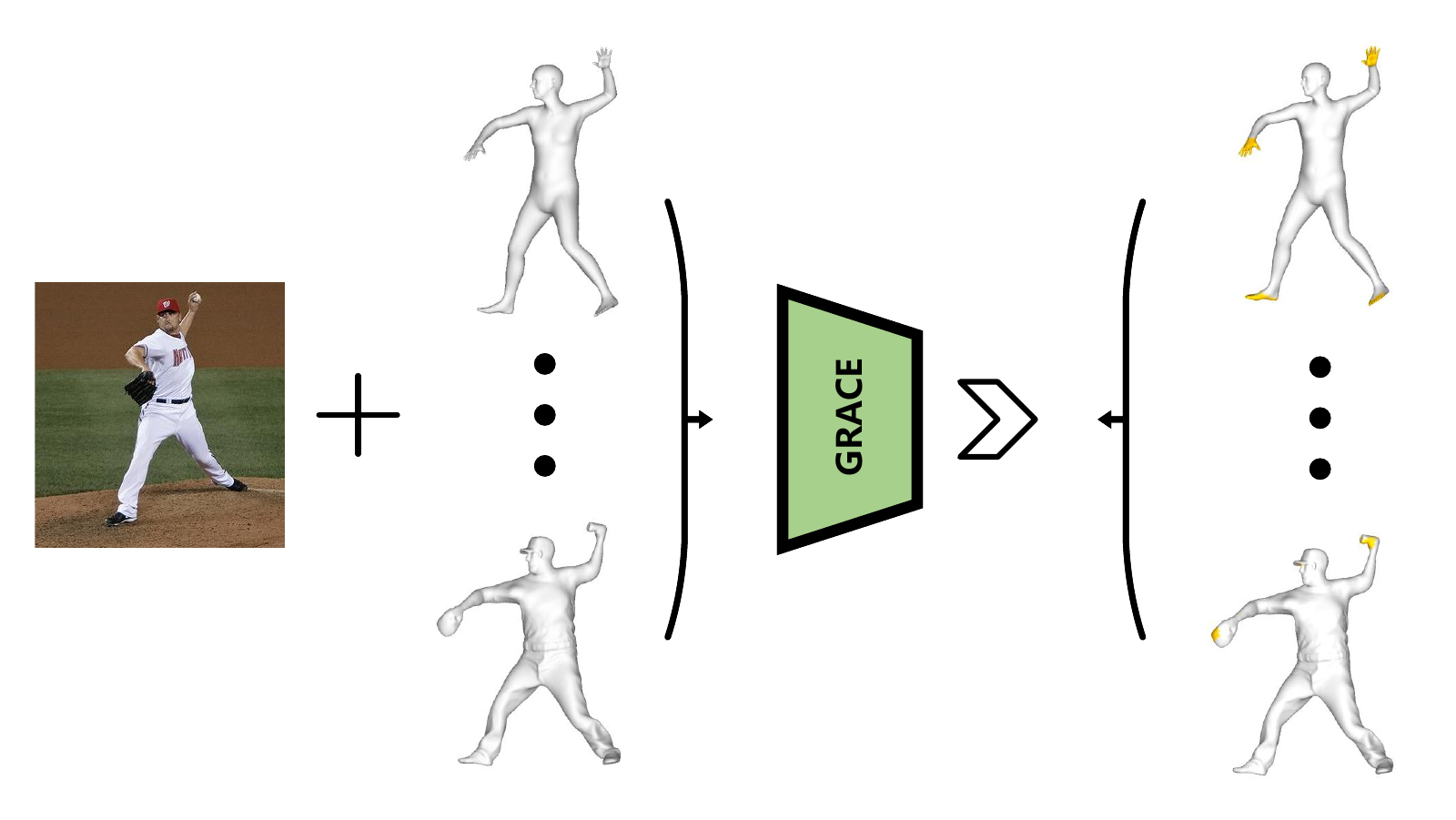}
          \caption{Given a monocular image and its paired arbitrary 3D human point cloud, GRACE accurately predicts human-scene contact. The yellow-highlighted regions indicate the predicted dense contact areas.}
          \label{fig:first}
	\vspace{-0.5cm}
\end{figure}
% \begin{figure}[t]
%   \centering
%   \small
%         \begin{overpic}[width=\linewidth]{figs/first.pdf}

%   \end{overpic}
%   \caption{Given a monocular image and its paired arbitrary 3D human point cloud, GRACE accurately predicts human-scene contact. The yellow-highlighted regions indicate the predicted dense contact areas.}
%   \label{fig:first}
% \end{figure}

\section{Introduction}
\label{sec:intro}

% \begin{figure}[t]
% 	\centering
% 	\small
% 		\begin{overpic}[width=0.92\linewidth]{figs/first.pdf}
		    
% 	\end{overpic}
% 	\caption{\textbf{Grounding Affordance from Interactions.} We propose to ground 3D object affordance through 2D interactions. Given an object point cloud with an interactive image, grounding the corresponding affordance on the 3D object.}
%  \label{fig1}
% \end{figure}

% Understanding human-scene interactions is a fundamental challenge in computer vision. Beyond recognizing human pose and action, it is crucial to infer the physical contact relationships between the human body and the surrounding environment. In 3D human behavior analysis and scene interaction modeling, dense contact estimation serves as a critical bridge between the virtual and physical worlds. Accurately predicting contact regions between human body surface vertices and scene objects enables physically plausible pose optimization \cite{nam2024contho} (e.g., reducing foot-sliding and body-object penetration), motion generation \cite{kulkarni2024nifty,xu2023interdiff}, and scene layout reasoning, thereby providing essential support for applications in robotics \cite{jiqir1,mandikal2021graff}, AR/VR \cite{cheng2013affordances}, and reconstruction \cite{xie2022chore,xu2021d3d}.

Understanding human-scene interactions is a fundamental challenge in computer vision. Beyond recognizing human pose and action, it is essential to infer the physical contact relationships between the human body and its surrounding environment. In 3D human behavior analysis and scene interaction modeling, dense contact estimation serves as a critical bridge between the virtual and physical worlds. Accurately predicting contact regions between human body surface vertices and scene objects enables physically plausible pose optimization \cite{nam2024contho} (e.g., reducing foot-sliding and body-object penetration), motion generation \cite{kulkarni2024nifty,xu2023interdiff}, and scene layout reasoning, thereby supporting applications in robotics \cite{jiqir1,mandikal2021graff}, AR/VR \cite{cheng2013affordances}, and 3D reconstruction \cite{xie2022chore,xu2021d3d}.

% Recent approaches have focused on predicting dense 3D contact on human body vertices from monocular images. Huang et al. \cite{Huang:CVPR:2022}  introduced a dataset with paired images, 3D human bodies, and scene meshes, leveraging a transformer-based architecture to infer 3D contact on a parametric human model. DECO \cite{tripathi2023deco} extended this paradigm to in-the-wild images by introducing the DAMON dataset and designing a dual-branch network that integrates scene context and body part features for contact prediction. However, these methods predominantly rely on parametric human models (e.g., SMPL \cite{SMPL:2015}) and establish correspondences between image features and contact regions via predefined vertex sequence mappings. While effective on structured data, their performance is inherently constrained by the fixed ordering of mesh vertices. When applied to unstructured human point clouds reconstructed from monocular images, the randomization of vertex indices and spatial distribution disrupts sequence mapping, leading to geometrically inconsistent contact predictions. Alternative approaches such as LEMON \cite{yang2023lemon}, IAG \cite{Yang_2023_ICCV}, and CONTHO \cite{nam2024contho} integrate 2D images and 3D point clouds for interaction reasoning but suffer from dependencies on precise 3D object models.

Recent methods have focused on estimating dense 3D contact on human body vertices from monocular images. Huang et al. \cite{Huang:CVPR:2022} introduced a dataset with paired RGB images, 3D human bodies, and scene meshes, and leveraged a transformer-based architecture to infer 3D contact on parametric human models. DECO \cite{tripathi2023deco} extended this paradigm to in-the-wild imagery by releasing the DAMON dataset and proposing a dual-branch network that integrates scene context and body part features for contact prediction. However, these methods predominantly rely on parametric human models (e.g., SMPL \cite{SMPL:2015}) and establish correspondences between image features and contact regions via predefined vertex sequence mappings. While effective on structured meshes, their performance is fundamentally constrained by the fixed vertex ordering. When applied to unstructured human point clouds reconstructed from monocular images, the lack of consistent indexing and irregular spatial distribution disrupts this sequence-based mapping, leading to geometrically inconsistent contact predictions. Alternative approaches such as LEMON \cite{yang2023lemon}, IAG \cite{Yang_2023_ICCV}, and CONTHO \cite{nam2024contho} incorporate both 2D images and 3D point clouds for interaction reasoning but often rely on accurate object models, which limits their applicability.

In this paper, we propose incorporating 3D human point clouds with corresponding image features, leveraging both 2D interaction semantics and 3D geometric priors for 3D dense contact estimation. In monocular images, human-scene contact is often occluded or incomplete, which inherently restricts direct visual cues available for inference. By incorporating human geometric priors, our model capitalizes on spatial consistency constraints—including topological relationships and articulated structure—to ensure physically plausible contact estimation even under severe occlusions or incomplete observations. Crucially, rather than directly mapping features to predefined vertex sequences of the human mesh, our approach learns an implicit mapping between geometric features and contact probabilities. This novel paradigm accommodates arbitrary human point clouds regardless of their adherence to SMPL topology, thus enabling generalization to diverse 3D human representations, including non-SMPL point clouds (Sec. \ref{sec 4.4:exp.Extra experiment}).

To achieve this goal, we propose the \textbf{GRACE}, which effectively integrates 2D interaction semantics with 3D human geometric priors to enable robust, accurate, and generalizable contact prediction. In detail, it contains two carefully designed sub-modules: one is the \textbf{H}ierarchical \textbf{F}eature \textbf{E}xtraction \textbf{M}odule (\textbf{HFEM}) that extracts informative multi-level features from both images and corresponding point clouds, and the other is the \textbf{M}ulti-level \textbf{F}eature \textbf{F}usion \textbf{M}odule (\textbf{MFFM}) that fuses the extracted multi-level features from the two modalities to obtain a comprehensive point cloud representation that is enriched with image interaction semantics. Finally, the fused representation is processed by a point cloud decoder, symmetrically designed with respect to the point cloud encoder, which directly regresses dense per-vertex contact probabilities without relying on any predefined mesh topology or fixed vertex order. This paradigm allows GRACE to move beyond the limitations of conventional fixed-sequence mapping methods and to be directly applied to unstructured or arbitrary human point clouds. By integrating 3D human geometric priors (such as topological consistency and kinematic constraints) GRACE maintains physical plausibility even under severe occlusions or partial observations. For example, when visual evidence of seated postures is missing, geometric cues (e.g., pelvis curvature and thigh orientations) can effectively guide the inference of chair-seat contact regions.

The contributions are summarized as follows:
% \begin{itemize}[leftmargin=15pt,topsep=0pt,itemsep=2pt]
\begin{itemize}
    \item[\textbf{1)}]We propose a novel paradigm for 3D dense contact estimation, which extends contact reasoning to unstructured human point clouds by learning an implicit mapping between geometric features and contact probabilities. This provides a scalable foundation for future research.
    \item[\textbf{2)}] We introduce the GRACE framework, which integrates image-based contact cues with 3D human priors through a hierarchical cross-modal reasoning mechanism, enabling more robust and generalizable contact estimation.
    \item[\textbf{3)}] We refine geometric error metrics to provide a more comprehensive evaluation of contact prediction quality. Our method achieves state-of-the-art performance on HSI benchmark. Moreover, it surpasses existing HSI approaches on the HOI dataset, demonstrating the superiority of GRACE.
\end{itemize}

\section{Related Work}

\subsection{Human-Scene Interaction in 2D}
Current approaches for representing 2D Human-Object Interaction (HOI) and Human-Scene Interaction (HSI) exhibit considerable diversity. Early-stage 2D HOI methods mainly concentrate on localizing humans and objects via bounding boxes while establishing semantic interaction associations. Representative approaches such as \cite{kim2021hotr,Qi2018LearningHI,deepcon2019,learningHOI2017,zou2021_hoitrans} perform entity localization and interaction categorization through action-type classification. However, these methods predominantly rely on coarse semantic labels and lack the capability for contact inference. HOT \cite{chen2023hot} introduces 2D contact annotations to facilitate human-object contact estimation in images. These efforts have achieved notable progress in understanding human-object interaction relationships within 2D space. Nevertheless, they still encounter significant challenges when extending to 3D spatial applications due to dimensional limitations.

\subsection{Human-Scene Interaction in 3D}
Three-dimensional contact information plays a critical role in multiple domains including 3D human pose estimation \cite{RempeContactDynamics2020,PhysCapTOG2020,xie2022chore}, 3D hand pose estimation \cite{Cao2020ReconstructingHI,grady2021contactopt,hasson19_obman}, 3D human motion generation \cite{rempe2021humor,taheri2021goal,Zhang:ICCV:2021}, and 3D scene layout estimation \cite{yi2022mime}. While recent methods \cite{NEURIPS2021_REMIPS,scene_track2000,PhysCapTOG2020,Mueller_self_con2021,fieraru2021learning} have achieved notable progress by predicting joint-level and patch-level contacts, these sparse contact annotations fundamentally fail to capture the intricate nature of full-body environmental interactions. The coarse discretization of contact regions inadequately represents the continuous pressure distribution and multi-scale contact patterns inherent in human-world engagements.

Dense human contact estimation has been predominantly explored through SMPL-based parametric models \cite{SMPL:2015,SMPL-X:2019,MANO:SIGGRAPHASIA:2017}, with extensions to diverse tasks \cite{Hassan:CVPR:2021,huang2023diffusion,huang2022intercap,yin2023rotatingseeinginhanddexterity}. Early works like POSA \cite{Hassan:CVPR:2021}  learn body-centric priors to predict potential contact vertices conditioned solely on 3D human poses, while ignoring critical visual evidence from images. BSTRO \cite{Huang:CVPR:2022} leverages the transformer architecture \cite{devlin-etal-2019-bert} to infer contacts from monocular images but lacks dedicated network designs for human-scene contact estimation. DECO \cite{tripathi2023deco} introduces a dual-branch architecture exploiting scene context and body-part context, coupled with the DAMON dataset of natural HSIs. However, the model learns a mapping from the interaction semantics in the image to the human vertex sequence and rarely utilizes geometric information, thus failing to adapt to reasoning on unordered human point clouds. CONTHO \cite{nam2024contho} reconstructs 3D human bodies and objects from images, aligning image features with 3D vertex features through perspective projection and predicting contact based on human feature interactions. However, its generalization is limited, as it struggles in natural scenes and cannot generalize to a wider variety of real-world object types.\begin{figure*}[t]
	\centering
    \scriptsize
	\begin{overpic}[width=1.\linewidth]{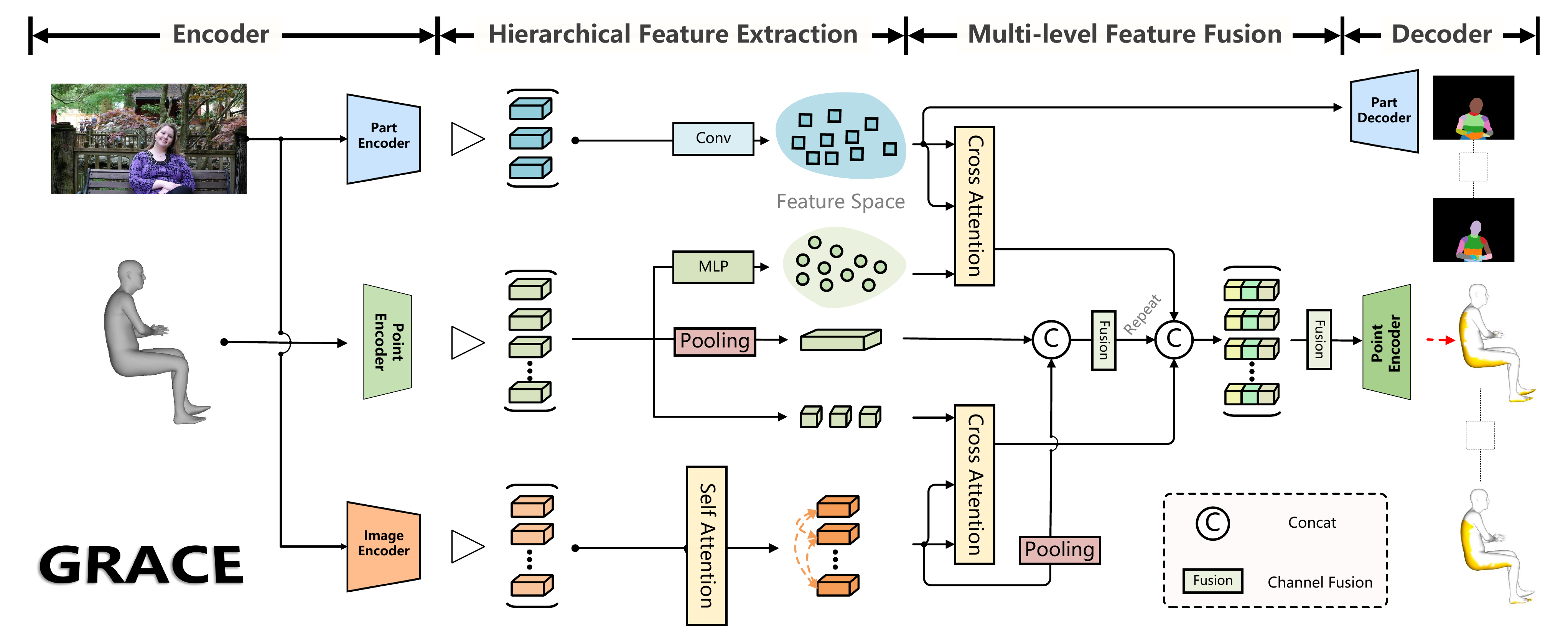}
     %\put(19,24){\scriptsize$P$}
        \put(9.2,27){${\mathbf{I}}$}
        \put(9.5,12.3){${\mathbf{P}}$}
        \put(33,36){$\mathbf{F}_p$}
        \put(33.3,10.7){$\mathbf{F}_i$}
        \put(33.2,24.3){$\mathbf{F}_h$}
        \put(52.7,10.2){\scriptsize$\hat{\mathbf{F}}_{i}$}
        \put(52.7,36){\scriptsize$\hat{\mathbf{F}}_{p}$}
        \put(48.3,25.5){$\mathbf{F}_{hp}$}
        \put(48.3,20.5){$\mathbf{F}_{hg}$}
        \put(59.3,24.5){\scriptsize $\mathbf{Q}_1$}
        \put(59.3,28.5){\scriptsize $\mathbf{K}_1$}
        \put(59.3,32.5){\scriptsize $\mathbf{V}_1$}
        \put(59.2,14.9){\scriptsize $\mathbf{Q}_2$}
        \put(59.3,10.8){\scriptsize $\mathbf{K}_2$}
        \put(59.3,6.9){\scriptsize $\mathbf{V}_2$}
        \put(63.8,25.6){\scriptsize $\mathbf{\Theta}_{1}$}
        \put(63.8,13.2){\scriptsize $\mathbf{\Theta}_{2}$}
        \put(67.2,7.4){\scriptsize $\mathbf{F}_{ig}$}
        \put(68,20.4){\scriptsize $\mathbf{F}_{g}$}
        \put(79,24.7){\scriptsize$\mathbf{F}_{c}$}
        \put(85.2,20.2){\scriptsize$\hat{\mathbf{F}}_{c}$}
        \put(91,20.5){\scriptsize $\hat{y}$}
        \put(93.5,12.9){\scriptsize $\mathbf{\mathcal{L}}_{c}$}
        \put(93,30.1){\scriptsize $\mathbf{\mathcal{L}}_{p}$}
        \put(97.4,33.8){\scriptsize $\hat{\mathbf{M}}_p$}
        \put(97.4,25.7){\scriptsize $\mathbf{M}_p$}
        
        % \put(70,21.5){\scriptsize $\mathbf{K}_1$}
        % \put(70.2,19){\scriptsize $\mathbf{V}_1$}
        % \put(70.4,2.6){\scriptsize $\mathbf{K}_2$}
        % \put(70.2,5.2){\scriptsize $\mathbf{V}_2$}
        % \put(28,13.8){$f_i$}

        % \put(18,24.1){$\mathbf{F_{I}}$}
        
        % \put(18,7.5){$\mathbf{F}_p$}
        % \put(1.7,11.3){\tiny$B_{obj}$}
        % \put(6.3,11.3){\tiny$B_{sub}$}
        % \put(11.3,11.2){\tiny$M_{sce}$}
        % \put(60,18.6){\scriptsize$\mathbf{F}_{j}$}
        % \put(67.5,17.8){\scriptsize$\mathbf{F}_{s}$}
        % \put(67.5,6.5){\scriptsize$\mathbf{F}_{e}$}
        % \put(77.9,6){\scriptsize$\mathbf{F}_{\alpha}$}
        % \put(77.9,20.5){\scriptsize$\mathbf{F}_{i\alpha}$}
        % \put(94.5,20){\scriptsize$\mathbf{F}_{p\alpha}$}
        
        % \put(94.6,2.8){\scriptsize$\hat{\mathbf{F}}_{p}$}  

        % \put(90.5,16){\scriptsize$\Gamma$}
        % \put(92,13){\scriptsize$f_{\phi}$}
        
        % \put(96,16.5){\scriptsize $\hat{\phi}$}
        
    %figure_a
	\end{overpic}
	\caption{\textbf{Method.} Overview of Geometry-level Reasoning for 3D Human-scene Contact Estimation Network(GRACE), it first extracts image features $\mathbf{F}_{i},\mathbf{F}_{p}$ and point cloud features $\mathbf{F}_{h}$, then uses Hierarchical Feature Extraction Module (Sec. \ref{sec:3.2}) to extract $\hat{\mathbf{F}}_p$, $\hat{\mathbf{F}}_i$, $\mathbf{F}_{hp}$, $\mathbf{F}_{hg}$ and $\mathbf{F}_{ig}$. Next, Multi-level Feature Fusion (Sec. \ref{sec:3.3}) aligns $\mathbf{\Theta}_{1/2}$ and $\mathbf{F}_{g}$ to obtain the 3D contact fusion feature $\hat{\mathbf{F}}_{c}$. Finally, $\hat{\mathbf{F}}_{c}$ is sent to the point decoder to obtain the final contact prediction result $\hat{y}$.}
 \label{fig:method}
\end{figure*}
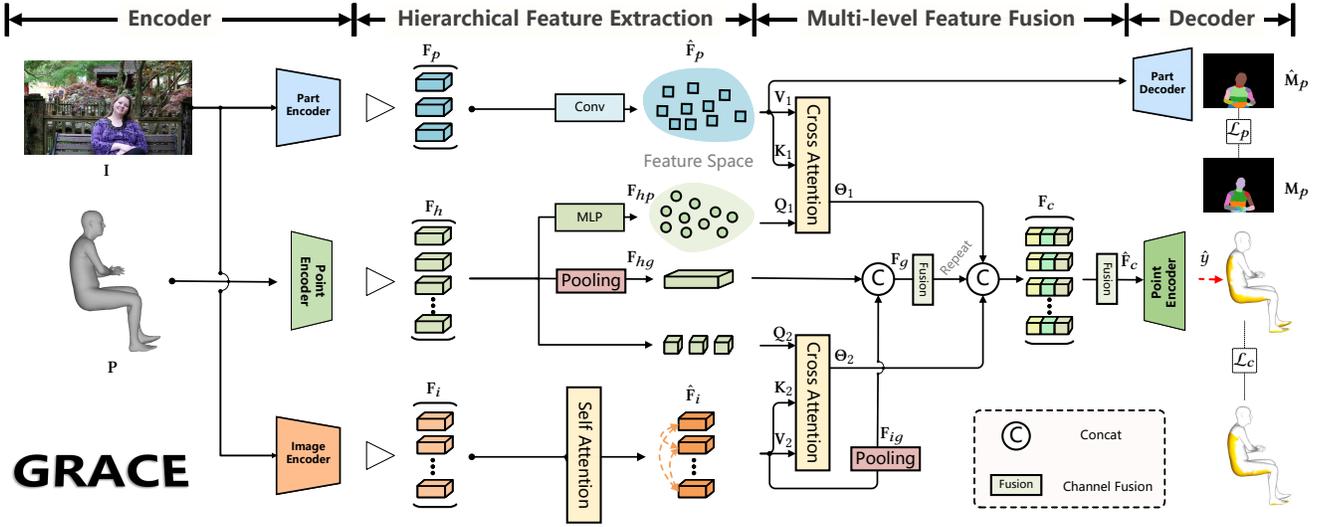 LEMON \cite{yang2023lemon} jointly process 2D images and 3D object point clouds but require precise object geometry inputs that limit practical applicability. Our framework integrates 2D interaction semantics with 3D human topological priors within a geometry-aware architecture, where visual embeddings directly guide the regression of contact probabilities on human point clouds. This approach eliminates the reliance on predefined human meshes, facilitating generalization to unstructured human point clouds.

\section{Method}
\label{sec:method}

\subsection{Overview}
\label{sec:3.1}
This paper aims to predict vertex-level contact regions on a three-dimensional human mesh from a single RGB image. Given the inputs $\{P,I\}$, where $P \in \mathbb{R}^{N \times 3}$ denotes the human point cloud, $N$ represents the total number of points in the point cloud. It is worth noting that in our training, we adopt vertices generated by the SMPL model \cite{SMPL:2015} as the point cloud input. The SMPL model parameterizes the human body using pose parameters $\theta \in \mathbb{R}^{72}$ and shape parameters $\beta \in \mathbb{R}^{10}$, and outputs a mesh $M(\theta,\beta) \in \mathbb{R}^{6890 \times 3}$. Meanwhile, $I \in \mathbb{R}^{3 \times H \times W}$ represents an RGB image, where $H$ and $W$ denote the image height and width.

As illustrated in Fig. \ref{fig:method}, the proposed \textbf{GRACE} network employs a multi-branch architecture. Specifically, it consists of two image encoders based on HRNet \cite{HRNet} and a three-dimensional point cloud encoder based on PointNeXt \cite{pointnext}. These encoders are responsible for extracting raw local features from the image and point cloud, denoted as $\mathbf{F}_i$, $\mathbf{F}_p \in \mathbb{R}^{C\times H'\times W'}$ and $\mathbf{F}_h \in \mathbb{R}^{C\times N_p}$, respectively. Both the image and point cloud encoders are initialized with weights pre-trained on large-scale datasets such as ImageNet \cite{deng2009imagenet} and ShapeNet \cite{chang2015shapenet}, thereby leveraging robust feature representations.

Subsequently, a multi-level feature extraction module further processes these raw features to generate intermediate representations at different hierarchies, denoted as $\hat{\mathbf{F}}_p$, $\hat{\mathbf{F}}_i$, $\mathbf{F}_{hp}$, $\mathbf{F}_{hg}$ and $\mathbf{F}_{ig}$. To facilitate effective cross-modal fusion, we design a fusion module that employs a cross-attention mechanism and concatenation operations to deeply integrate multi-level image semantic features and human geometric information, ultimately producing fused features $\hat{F}_c$ rich in interaction cues. Finally, $\hat{\mathbf{F}}_c$ is fed into a point cloud decoder, which is symmetrically designed with respect to the point cloud encoder, to generate a dense vertex-wise 3D contact probability vector. The overall forward propagation of the network can be succinctly expressed as:
\begin{equation}
\small
\hat{y} = f_{\theta}(I, P),
 \label{equ:overview}
\end{equation}
where $f_{\theta}$ denotes our proposed network.

% The \textbf{GRACE} network(Fig. \ref{fig:method}) employs a dual-branch image encoder (HRNet \cite{HRNet}) and a 3D point cloud encoder (PointNeXt \cite{pointnext}). For 2D images, the global image branch generates holistic features $F_i$, while the human-part attention branch focuses on region-specific human interactions features $F_p$ by upsampling $F_p$ using part decoder.
% For 3D inputs, the encoder processes image-aligned 3D human point clouds to extract human geometry features $F_h$, which inherently encode spatial priors of potential human-scene contact regions through body-part geometric properties.Then, the CAIFE module takes $F_i$ and $F_p$ as input, capturing contact-related features $F_c$ from the image. The 3D-2D Fusion module combines $F_c$ and 3D geometric features $F_h$ to integrate image-guided contact information and 3D body priors. Finally, the fused features are processed by the point cloud decoder to predict dense contact regions on the 3D human mesh.The network is represented as:

\subsection{Hierarchical Feature Extraction Module}
\label{sec:3.2}
To accurately model the interaction regions between the human body and the surrounding scene, we develop a multi-level feature extraction module that further processes the raw features from both the point cloud and the image to extract part-level semantic information and global contextual cues, thereby laying a robust foundation for subsequent cross-modal fusion.

For the human point cloud, the raw geometric features $\mathbf{F}_h \in \mathbb{R}^{C \times N_p}$ obtained via the PointNeXt encoder contain abundant local shape details; however, their direct representation often fails to explicitly reflect the semantic structure of human body parts. To address this, we design a non-linear mapping module based on a multi-layer perceptron (MLP), which applies point-wise transformations to project the raw point cloud features into a pre-defined part semantic space. The final output dimensionality is set to the number of human body parts $J$. This mapping compels the network to implicitly learn the correspondence between vertices and body parts (e.g., assigning high response weights to hand vertices), thereby effectively capturing the local structural information and establishing a fine-grained semantic basis for cross-modal feature alignment.In the image branch, after extracting intermediate features $\mathbf{F}_{p}$ using HRNet, we apply a composite convolutional layer—comprising convolution, batch normalization, and ReLU activation—to project the original image part features into a representation that is consistent with the point cloud part semantic space. To reinforce the semantic localization capability of this branch, an upsampling decoder is designed to map these features back to the original resolution, which is then supervised by a pixel-wise cross-entropy loss between the predicted part mask $\hat{\mathbf{M}}_p$ and the ground-truth mask $\mathbf{M}_p$. This design enhances the capacity of the image branch to capture local details, ensuring that the part features are semantically coherent.The transformation from the raw features of the two modalities to the part semantic space features can be expressed as:
\begin{equation}
% \small
\hat{F}_p = f_p(F_p), \quad F_{hp} = f_h(F_h),
 \label{equ:part}
\end{equation}
where \(f_p\) denotes a convolution operation and \(f_h\) denotes a multilayer perceptron (MLP).

Moreover, to further model the scene-level interaction information, the image branch also incorporates a self-attention Transformer layer \cite{transformer}  to capture contextual cues from the raw feature maps, thereby generating a scene interaction feature $\hat{\mathbf{F}_{i}}$.The self-attention mechanism is computed as:
\begin{equation}
\alpha_{ij} = \text{softmax}\left(\frac{\mathbf{Q}_j \mathbf{K}_{j}^{T}}{\sqrt{d}}\right), \quad \hat{\mathbf{F}}_{i} = \sum_j \alpha_{ij} \mathbf{V}_j,
 \label{equ:sa}
\end{equation}
where $\mathbf{Q}$, $\mathbf{K}$, and $\mathbf{V}$ denote the query, key, and value vectors, and $d$ is the feature dimensionality. To further enforce semantic regularization, the scene interaction feature $\hat{\mathbf{F}}_{i}$ is spatially averaged to yield a global semantic vector $\mathbf{F}_{ig} \in \mathbb{R}^{1 \times C}$. Correspondingly, the raw point cloud features $\mathbf{F}_h$ are also aggregated by global pooling to obtain a global geometric context feature $\mathbf{F}_{hg} \in \mathbb{R}^{1 \times C}$. Together, these global features provide a semantic prior that compensates for missing local information (e.g., due to occlusion) during the subsequent cross-modal fusion.

In summary, our multi-level feature extraction module constructs a rich representation through three complementary paths. This multi-level, cross-modal feature construction strategy not only preserves essential local structural and semantic information but also integrates global contextual cues, thereby providing a comprehensive information basis for efficient cross-modal fusion and accurate 3D human-scene contact prediction.
%%%%

\subsection{Multi-Level Feature Fusion Module}
\label{sec:3.3}

After the multi-level feature extraction stage, we further design a multi-level feature fusion module that aims to achieve hierarchical fusion of features from local to global levels, thereby accurately capturing the interaction regions between the 3D human body and the scene. The module comprises three core stages: (1) cross-modal feature alignment at the vertex level, (2) hierarchical semantic enhancement based on human part segmentation, and (3) feature integration under global relational constraints.

% Specifically, we first perform vertex-level cross-attention fusion between the raw geometric features of the point cloud and the scene interaction features extracted from the image. In this process, the point cloud features serve as queries $\mathbf{Q}$, while the image scene features serve as keys $\mathbf{K}$ and values $\mathbf{V}$. Under the standard multi-head attention mechanism, the computation is expressed as:
Specifically, $\mathbf{F}_{hp}$ and $\mathbf{F}_{h}$ is projected to form the  query $\mathbf{Q}_{1}, \mathbf{Q}_{2} = \mathbf{F}_{hp}\mathbf{W}_{1}, \mathbf{F}_{h}\mathbf{W}_{2}$; $\hat{\mathbf{F}}_{p}$ and $\hat{\mathbf{F}}_{i}$ are projected to form different keys and values $\mathbf{K}_{1}, \mathbf{K}_{2} = \hat{\mathbf{F}}_{p}\mathbf{W}_{3}, \hat{\mathbf{F}}_{i}\mathbf{W}_{4}; \mathbf{V}_{1}, \mathbf{V}_{2} = \hat{\mathbf{F}}_{p}\mathbf{W}_{5}, \hat{\mathbf{F}}_{i}\mathbf{W}_{6}$, where $\mathbf{W}_{1 \sim 6}$ are projection weights. Then, these features are aggregated via a cross-attention mechanism to achieve cross-modal information fusion, expressed as:
\begin{equation}
    \mathbf{\Theta}_{1/2} = softmax(\mathbf{Q}_{1/2}^{T}\cdot\mathbf{K}_{1/2}/\sqrt{d})\cdot\mathbf{V}^{T}_{1/2},
\end{equation}
where $\mathbf{Q}_{1/2} \in \mathbb{R}^{d \times N_p}, \mathbf{K}_{1/2}, \mathbf{V}_{1/2} \in \mathbb{R}^{d \times N_i}$, $d$ is the dimension of projection. In this way, the network is able to learn the correspondence between each point and potential interaction regions in the global scene, thereby integrating image semantics and point cloud geometry at the vertex level. Meanwhile, this process also enables local geometric information of the human body (e.g., hands or torso) to "query" the corresponding regions in the image and retrieve supplementary semantics from the image part features $ \hat{\mathbf{F}}_{p} $, thus enhancing the fine-grained perception of potential contact areas.

\begin{table*}[t]
\centering
% \small
\caption{Comparison of GRACE with SOTA models on the HSI task datasets RICH \cite{Huang:CVPR:2022} and DAMON \cite{tripathi2023deco}. POSA$^{\text{GT}}$ means taking ground-truth bodies as input, while POSA$^{\text{PIXIE}}$ takes the estimated
 bodies from PIXIE \cite{feng2021collaborative}.} %See discussion in Sec. \ref{Sec:4.2 exp_comparison results}}
\label{tab:exp_hsi}
% \resizebox{\textwidth}{!}{
\begin{tabularx}{\textwidth}{lXXXX XXXX}

% \begin{tabular}{lcccc cccc}
\toprule
\multirow{2}[2]{4em}{\bf Methods} & 
\multicolumn{4}{c}{RICH \cite{Huang:CVPR:2022}} & 
\multicolumn{4}{c}{DAMON \cite{Huang:CVPR:2022}} \\
\cmidrule(lr){2-5} \cmidrule(lr){6-9}
& Pre$\uparrow$ & Rec$\uparrow$ & F1$\uparrow$ & Geo.sum(cm)$\downarrow$ & Pre$\uparrow$ & Rec$\uparrow$ & F1$\uparrow$ & Geo.sum(cm)$\downarrow$ \\
\midrule
POSA$^{\text{PIXIE}}$~\cite{PIXIE:2021, Hassan:CVPR:2021} & 0.31 & 0.69 & 0.39 & 58.4 & 0.42 & 0.34 & 0.31 & 73.5 \\ 
POSA$^{\text{GT}}$~\cite{PIXIE:2021, Hassan:CVPR:2021} & 0.37 &  0.76 & 0.46 & 56.2 & - & - & - & -  \\
BSTRO \cite{Huang:CVPR:2022} & 0.61 & 0.65 & 0.59 & 46.2 & 0.51 & 0.54 & 0.47 & 62.6 \\
DECO \cite{tripathi2023deco} & 0.61 & 0.63 & 0.58 & 49.6 & 0.65 & 0.57 & 0.55 & 51.2 \\  \hline
Ours & \textbf{0.63} & \textbf{0.77} & \textbf{0.66} & \textbf{19.2} & 0.61 & \textbf{0.65} & \textbf{0.58} & \textbf{38.0} \\
\bottomrule
% \end{tabular}
\end{tabularx}
% }
\end{table*}

\begin{table*}[t]
\centering
% \small
\caption{Comparison of GRACE on the HOI task datasets 3DIR \cite{yang2023lemon} and BEHAVE \cite{bhatnagar22behave}.}
\label{tab:exp_hoi}
% \resizebox{\textwidth}{!}{

\begin{tabularx}{\textwidth}{lXXXX XXXX}
% \begin{tabular}{lcccc cccc}
\toprule
\multirow{2}[2]{4em}{\bf Methods}& 
\multicolumn{4}{c}{3DIR \cite{yang2023lemon}} & 
\multicolumn{4}{c}{BEHAVE \cite{bhatnagar22behave}} \\
\cmidrule(lr){2-5} \cmidrule(lr){6-9}
& Pre$\uparrow$ & Rec$\uparrow$ & F1$\uparrow$ & Geo.sum(cm)$\downarrow$ & Pre$\uparrow$ & Rec$\uparrow$ & F1$\uparrow$ & Geo.sum(cm)$\downarrow$ \\
\midrule
BSTRO \cite{Huang:CVPR:2022} & 0.66 & 0.57 & 0.58 & 44.1 & 0.27 & 0.18 & 0.17 & 64.4 \\
DECO \cite{tripathi2023deco} & \textbf{0.68} & 0.64 & 0.62 & 41.2 & \textbf{0.57} & 0.23 & 0.16 & 71.9 \\\hline	
Ours & 0.57 & \textbf{0.75} & \textbf{0.63} & \textbf{24.5} & 0.29 & \textbf{0.53} & \textbf{0.25} & \textbf{61.1 }\\
\bottomrule
% \end{tabular}
% }
\end{tabularx}
\end{table*}
% BSTRO\cite{Huang:CVPR:2022}  0.263 & 0.171 & 0.169 & 64.4 \\
% DECO\cite{tripathi2023deco}  0.585 & 0.240 & 0.161 & 71.9 \\
% Ours & 0.570 & \textbf{0.752} & 0.625 & \textbf{24.5} & 0.282 & \textbf{0.528} & \textbf{0.249} & \textbf{61.0
% LEMON\cite{yang2023lemon}  & \textbf{0.73} & \textbf{0.76} & \textbf{0.73} & \textbf{21.1} & 0.31 & 0.42 & \textbf{0.29} & \textbf{58.3} \\ \hline	
% BSTRO \cite{Huang:CVPR:2022} & 0.66 & 0.57 & 0.58 & 44.1 & 0.27 & 0.24 & 0.21 & 66.5 \\
% DECO \cite{tripathi2023deco} & 0.68 & 0.64 & 0.62 & 41.2 & \textbf{0.58} & 0.23 & 0.22 & 64.7 \\\hline	
% Ours & 0.57 & \textbf{0.75} & \textbf{0.63} & \textbf{24.5} & 0.25 & \textbf{0.66} & \textbf{0.27} & \textbf{58.6 }\\
Furthermore, we integrate the global features extracted from both modalities to enrich the interaction representation. The global features from the image branch, obtained via spatial average pooling of the context-enhanced scene feature, and the global geometric context features from the point cloud branch are concatenated and further refined through channel fusion operations (a lightweight MLP). The resulting global representation is then concatenated with the fusion outputs from the aforementioned cross-attention stages, which not only provides a macroscopic constraint for local interaction but also serves as a compensatory cue in cases of occlusion or missing local information, thus ensuring the overall consistency and robustness of the prediction. The above process can be expressed as:
\begin{equation}
\hat{\mathbf{F}}_c = f_{\xi}\left( \left[ \Theta_1,\, \Theta_2,\, f_{r}(f_{\xi}(\mathbf{F}_g)) \right] \right),
 \label{equ:CA}
\end{equation}
where  $f_{\xi}$ denotes the channel-wise fusion operator ,  $f_{r}$ refers to the repeat operation that expands the global feature $ \mathbf{F}_g \in \mathbb{R}^{1 \times C} $ along the spatial dimension, and $[\cdot]$ represents the concatenation operation.

In summary, the vertex-level cross-attention enables the network to focus on the mapping between local geometry and scene context, the part-level cross-attention guides the interaction between anatomical priors and local image details, and the global feature fusion provides an overarching scene constraint. This multi-faceted fusion mechanism ensures that the network captures both detailed local contact cues and maintains comprehensive global interaction awareness.

% \begin{equation}
% F_{f} = f_{\gamma}(\text{ReLU}(\text{LayerNorm}(f[\mathbf{F}_{hc},\mathbf{F}_{hs}])),
% \label{equ:3dfinal}
% \end{equation}

% where  $[\cdot]$ denotes the concatenation, $f \in \mathbb{R}^{2D \times D} $ is a learnable projection matrix, LayerNorm represents layer normalization, \( f_{\gamma} \) is the final SA Transformer refinement function, and \( \theta \) denotes the learnable parameters of the model.

\subsection{Decoder and Loss Functions}
\label{sec:3.4}
The Part Decoder upsamples the image part features to generate a part segmentation mask, which is subsequently supervised by the ground-truth segmentation mask to ensure an accurate representation of the 2D spatial distribution of human body parts.The Point Cloud Decoder decodes the final features $\hat{\mathbf{F}}_{c}$ through a Feature Propagation layer to regress the predicted contact results on the vertices, denoted as $\hat{y} \in \mathbb{R}^{N \times C_{1}}$. These results are then mapped via a multi-layer perceptron (MLP) to $N \times 1$, obtaining the dense contact prediction on the 3D human mesh vertices. Thanks to the integration of 3D human point cloud information and the point cloud encoder-decoder network architecture, \textbf{GRACE} overcomes the limitations of previous methods that were restricted to sequence mappings, enabling contact prediction for both ordered and unordered human models (Sec. \ref{sec 4.4:exp.Extra experiment}).

The overall training loss is expressed as:
\begin{equation}
\small
\mathbf{\mathcal{L}_{total}} = \omega_{1}\mathbf{\mathcal{L}}_{c} + \omega_{2}\mathbf{\mathcal{L}}_{p},
\end{equation}
where $\mathcal{L}_{p}$ represents the segmentation loss between the predicted and the ground truth part segmentation mask, and $\mathcal{L}_{c}$ represents the loss between the predicted contact and the ground truth contact, which is a combination of Focal loss \cite{lin2017focal} and Dice loss \cite{milletari2016v}.

Contact regions typically cover only a small portion of the 3D human body surface (e.g., the foot-ground contact or hand-object interaction), leading to an extreme class imbalance between positive and negative samples (contact points vs non-contact points). Focal loss addresses this global imbalance through sample-level weighting, while Dice loss enhances local sensitivity through region-level optimization. These two losses complement each other, significantly improving the recall rate and localization accuracy of sparse contact points.
\section{Experiment}
\label{sec:exp}

\begin{figure*}%
\centering %
\includegraphics[trim=000 000 000 000,clip,width=\linewidth]{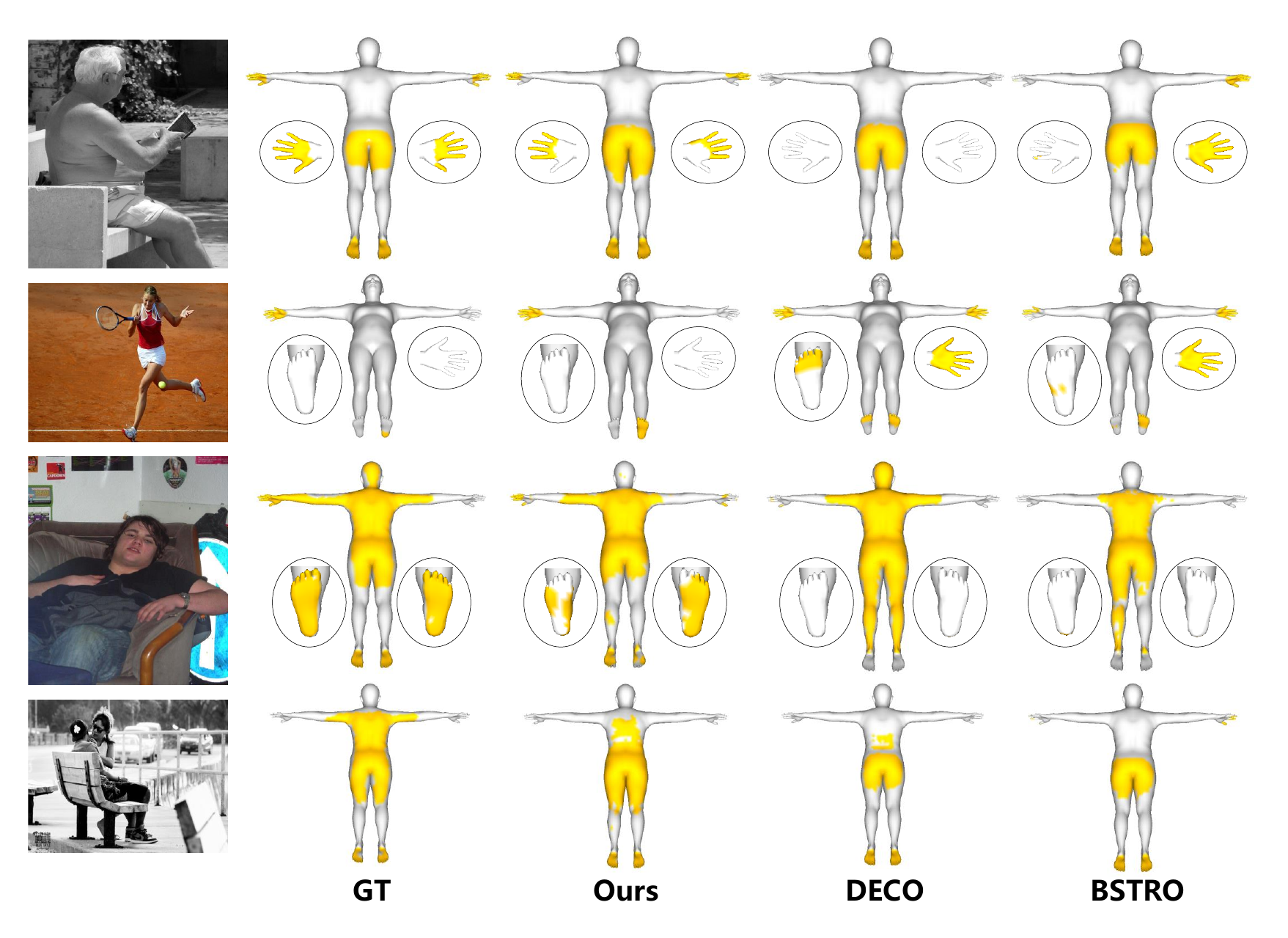}
\vspace{-1.5 em}
\caption{\textbf{Visualization Results.} Qualitative evaluation of GRACE, DECO \cite{tripathi2023deco} and BSTRO \cite{Huang:CVPR:2022}  , alongside Ground Truth. The \textcolor[rgb]{0.88,0.82,0.36}{yellow} regions indicate the areas of contact between the human body and the scene.%
}
\label{fig:comparison_main}
\end{figure*}
\subsection{Benchmark Setting}
\label{Sec:exp.Benchmark Setting}
\noindent\textbf{Implementation Details.\ }For the feature extraction, we utilize HRNet \cite{HRNet} and PointNeXt \cite{pointnext} as the encoders for image and point cloud data, respectively, and load their pre-trained weights. To obtain ground-truth part segmentations, denoted as ${M}_p \in \mathbb{R}^{H \times W \times (J+1)}$, we follow the approach of \cite{tripathi2023deco,Kocabas_PARE_2021} to segment the posed ground-truth mesh  into $J = 24$ parts and render each part mask as a separate channel.

\noindent\textbf{Training and Evaluation.\ }We conduct extensive comparative experiments on the DAMON \cite{tripathi2023deco} and RICH \cite{Huang:CVPR:2022} datasets for the Human-Scene Interaction (HSI) task. Additionally, we evaluate our model's performance on the Human-Object Interaction (HOI) task using the BEHAVE \cite{bhatnagar22behave} and 3DIR \cite{yang2023lemon} datasets. For each dataset, we train our model independently on the training set and evaluate it on the corresponding test set. 

\noindent\textbf{Evaluation Metrics.\ }To assess the estimated dense Human-Scene Contact, we adopt standard detection metrics, including \textbf{pre}cision, \textbf{rec}all, and \textbf{F1} score, consistent with previous methods \cite{Huang:CVPR:2022,tripathi2023deco}. In prior works \cite{Huang:CVPR:2022,tripathi2023deco}, the geometric error metric (geo.err) has primarily considered the geometric error associated with false positives, while neglecting the error stemming from false negatives. This limited evaluation metric does not fully reflect the model's performance, particularly in terms of recall. To overcome this limitation, we propose a more comprehensive evaluation metric, termed Total Geometric Error (\textbf{geo.sum}). This metric accounts for both false positive and false negative geometric errors, providing a more balanced and thorough assessment of the accuracy of contact predictions. Further details on the implementation and training can be found in the supplementary material.

\begin{table}[t]
\centering
 % 缩小字体
\renewcommand{\arraystretch}{1.3}
\caption{\textbf{Ablation Study.} Performance when not introducing global feature (global.), part feature (part.), both global and part feature( p.\&g.), combination loss ($\mathbf{\mathcal{L}}_{c}$). \ding{55} means without. Bold denotes best performance.}
\label{tab:exp_ablation}
% \resizebox{\columnwidth}{!}{%
% \small
\begin{tabularx}{\columnwidth}{lXXXX XXXX}
% \begin{tabular}{@{}lrrrr@{}} % @{}去除两侧边距，r对齐优化数字列
\toprule
\textbf{Variant} & \textbf{Pre $\uparrow$} & \textbf{Rec $\uparrow$} & \textbf{F1 $\uparrow$} & \textbf{Geo.sum$\downarrow$} \\
\midrule
\cellcolor{mygray}\textbf{Ours} & \cellcolor{mygray}\textbf{0.609} & \cellcolor{mygray}\textbf{0.652} & \cellcolor{mygray}\textbf{0.581} & \cellcolor{mygray}\textbf{38.0} \\
\hline
\ding{55}  $\mathbf{\mathcal{L}}_{c}$ & 0.602 & 0.615 & 0.568 & 39.2 \\
\ding{55} part. & 0.573 & 0.651 & 0.559 & 40.8 \\
\ding{55} global. & 0.589 & 0.629 & 0.563 & 38.7 \\
\ding{55} p.\&g. & 0.584 & 0.626 & 0.556 & 41.1 \\
\bottomrule
% \end{tabular}
% }
\end{tabularx}
\end{table}
\subsection{Comparison Results}
\label{Sec:4.2 exp_comparison results}
We conducted a comprehensive comparison between our method and state-of-the-art approaches for the HSI task across multiple datasets, evaluating performance on both HSI and HOI tasks. As shown in Tab. \ref{tab:exp_hsi}, our method outperforms all other approaches on the RICH and DAMON datasets in terms of both F1 score and total geometric error (Geo.sum), demonstrating its superior predictive capability for the HSI task.
\begin{figure}
    \centering

    \begin{overpic}[width=1.\linewidth]{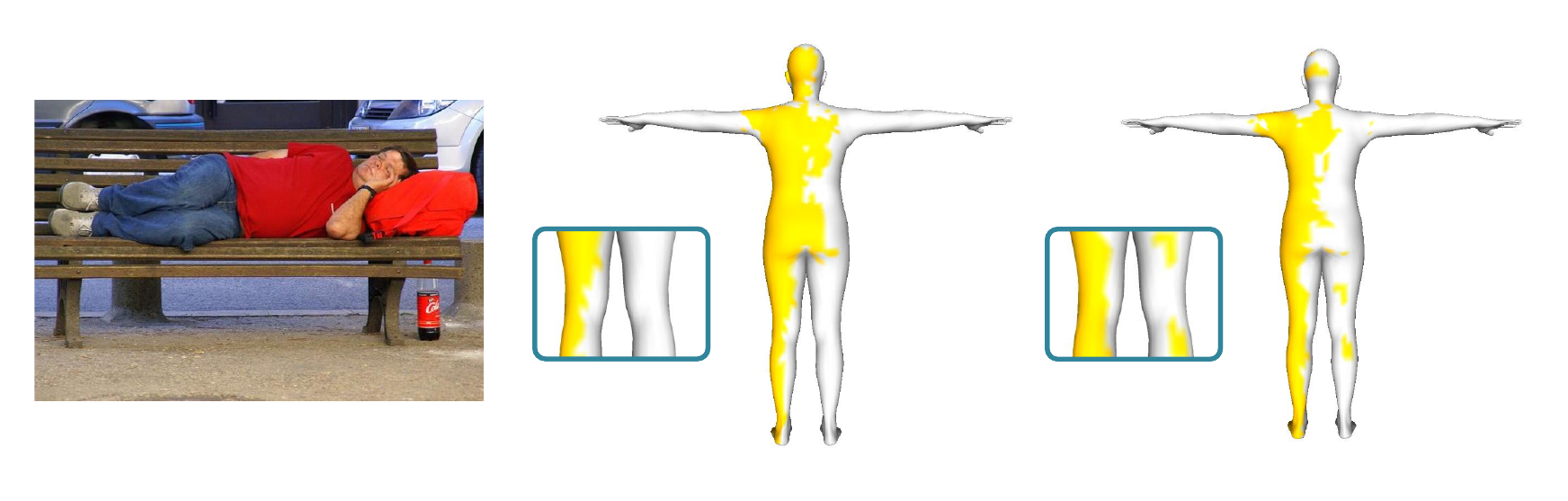}
    \put(46,-2){\textbf{$\bm{w }$ part.}}

    \put(79,-2){\textbf{$\bm{w/o }$ part.}}

    % \put(8.5,36.5){\textbf{\footnotesize Pour}}
    % \put(7.3,13){\textbf{\footnotesize Bottle}}
    % \put(75,-4){\textbf{(b)}}
    % \put(25,-4){\textbf{(a)}}

    \end{overpic}

    \caption{\textbf{Ablation of part feature branch. The result of contact estimation with ($\bm{w }$), without($\bm{w/o }$) part feature branch of image and point cloud.} }
    \label{fig:ab_part}
    \vspace{-10pt}
\end{figure}
\begin{figure}
    \centering
    \begin{overpic}[width=1.\linewidth]{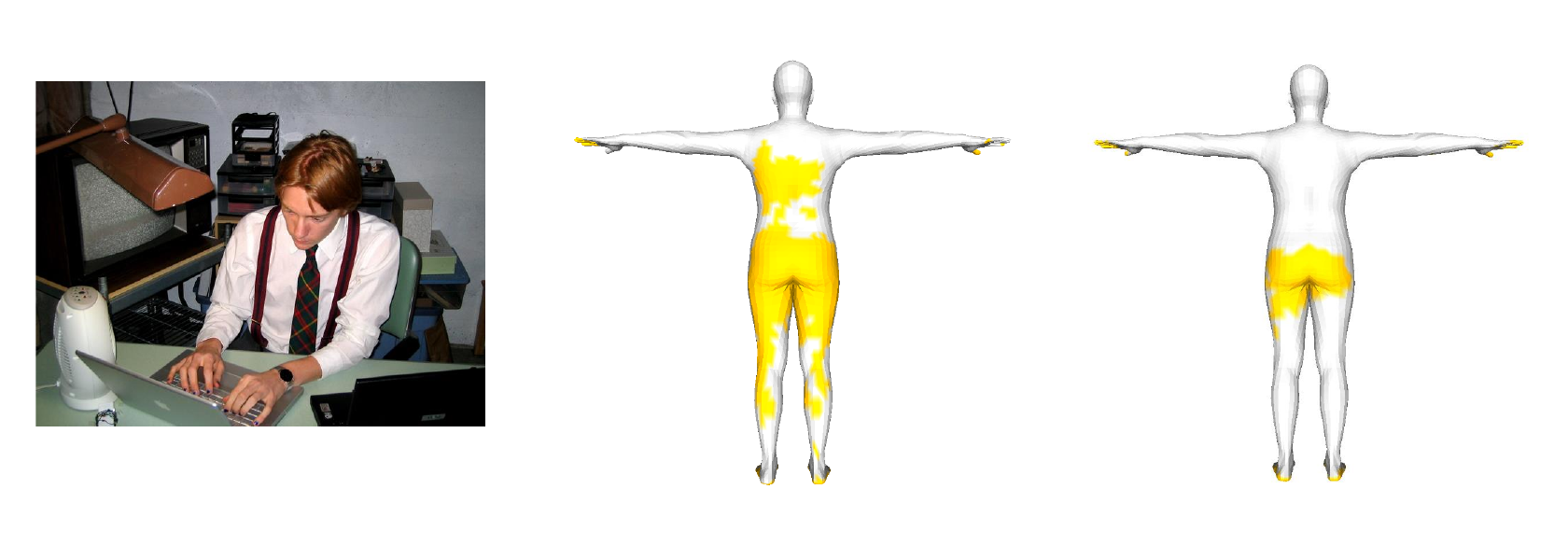}
    \put(46,-2){\textbf{$\bm{w }$ glo.}}

    \put(79,-2){\textbf{$\bm{w/o }$ glo.}}

    % \put(8.5,36.5){\textbf{\footnotesize Pour}}
    % \put(7.3,13){\textbf{\footnotesize Bottle}}
    % \put(75,-4){\textbf{(b)}}
    % \put(25,-4){\textbf{(a)}}

    \end{overpic}

    \caption{\textbf{Ablation of global feature branch.} The result of contact estimation with ($\bm{w }$), without($\bm{w/o }$) global feature branch of image and point cloud.}
    \label{fig:ab_glo}
    \vspace{-10pt}
\end{figure}
\begin{figure*}%
\centering %
\includegraphics[trim=000 000 000 000,clip,width=\linewidth]{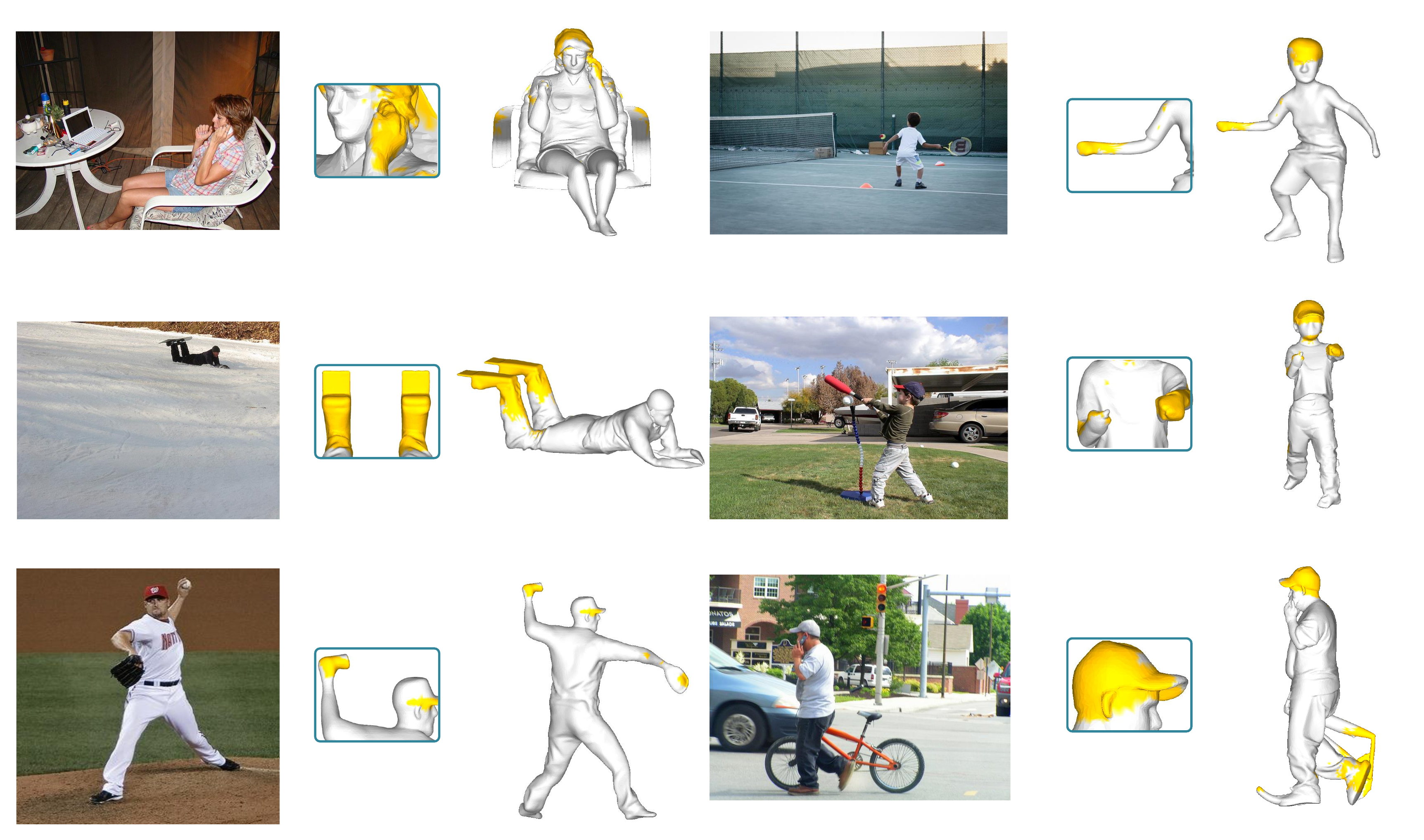}
\vspace{-1.5 em}
\caption{Contact prediction on unstructured human point clouds. Using Hunyuan 3D \cite{hunyuan3d22025tencent} to infer point clouds from monocular images (which lack a predefined topology and exhibit varying density and spatial distribution)}
\label{fig:hunyuan}
\end{figure*}
In Tab. \ref{tab:exp_hoi}, the performance on the HOI task varies across datasets. On the BEHAVE dataset—which primarily consists of indoor, human-made scenes with limited multi-object interactions—our method achieves the best results in both F1 score and Geo.sum. In contrast, the 3DIR dataset contains natural outdoor scenes where multi-object interactions are more common. Although both BEHAVE and 3DIR provide annotations for single-object contacts, the prevalence of multiple-object contact scenarios in 3DIR turns our strength in modeling complex interactions into a disadvantage. Since our model is not specifically designed to identify individual target objects, it often predicts contact between the human body and multiple surrounding objects, which leads to a drop in precision and thus affects the overall F1 score.In comparison, DECO adopts a more conservative prediction strategy, favoring “no contact” outputs. This results in higher precision but lower recall, as reflected in the experimental results. This explains why our model’s F1 score on the 3DIR dataset is only slightly higher than DECO’s. Nonetheless, our method still demonstrates a significant advantage in geometric consistency, achieving a Geo.sum of 24.5 compared to DECO’s 41.2, further validating the effectiveness and superiority of our approach in modeling human–object contact.

\subsection{Ablation Study} 
\label{sec:exp.ablation}
\begin{table}[t]
\footnotesize
\centering
  \renewcommand{\arraystretch}{1.2}
  \renewcommand{\tabcolsep}{6.pt}
   \caption{HPS estimation performance using contact derived from different sources. \ding{55} means without. }
    \label{tab:exp_HPS}
\resizebox{\columnwidth}{!}{%
\begin{tabular}{c|cccccc}
\toprule
  \textbf{Methods}  & \ding{55} \textbf{Contact} & \textbf{Prox} & \textbf{HOT} & \textbf{DECO} & \textbf{GRACE} & \textbf{GT}\\ \midrule
\textbf{V2V $\downarrow$} & 183.3 & 174.0 & 172.3 & 171.6 & \textbf{171.1} & 163.0 \\
\bottomrule
\end{tabular}
}
\end{table}

In Tab. \ref{tab:exp_ablation} and Fig. \ref{fig:ab_part} and Fig. \ref{fig:ab_glo}, we present the results of ablation experiments conducted on the core modules and loss functions of our model to quantitatively evaluate the contribution of each component to overall performance. The experimental results indicate that replacing the combined loss function (Focal Loss + Dice Loss) with BCE loss causes a simultaneous drop in both recall and F1 score, thereby highlighting the necessity of the combined loss for optimizing sparse contact regions. Moreover, the removal of the local feature branch (part.) leads to a significant decrease in precision by 5.9\% (from 0.609 to 0.573), while the recall remains almost unchanged (0.652 to 0.651). This confirms that the module, through human joint-space feature mapping, achieves fine-grained local alignment that effectively mitigates local mispredictions 
in the context of discrete contact inference (see Fig. \ref{fig:ab_part}). Concurrently, the Geo.sum metric increases from 38.0 to 40.8, underscoring the critical role of part-level information in improving both prediction accuracy and geometric consistency. In addition, removal of the global feature fusion branch (global.) results in a pronounced decrease in recall (from 0.652 to 0.629), indicating that global semantic information is indispensable for effectively reasoning about and compensating occluded regions (see Fig. \ref{fig:ab_glo}). 

Notably, when both the global and local feature branches are simultaneously removed, the model experiences a comprehensive degradation in performance (with the F1 score dropping to 0.556 and Geo.sum rising to 41.1, compared to 0.581 and 38.0 for the full model), a decline that exceeds the cumulative impact of removing each module individually. This finding reveals a synergistic effect between global constraints and local refinement, jointly ensuring the physical plausibility and spatial accuracy of the predictions. In summary, the multi-level fusion mechanism proposed in this paper—integrating global semantic guidance and local feature alignment—significantly enhances the precision and geometric rationality of contact prediction. Additionally, the adoption of a combined loss (Focal Loss + Dice Loss) proves particularly effective for optimizing sparse contact regions, thereby further contributing to the overall performance improvements demonstrated across all evaluated metrics.
\begin{figure}[t]
  \centering

        \begin{overpic}[width=\linewidth]{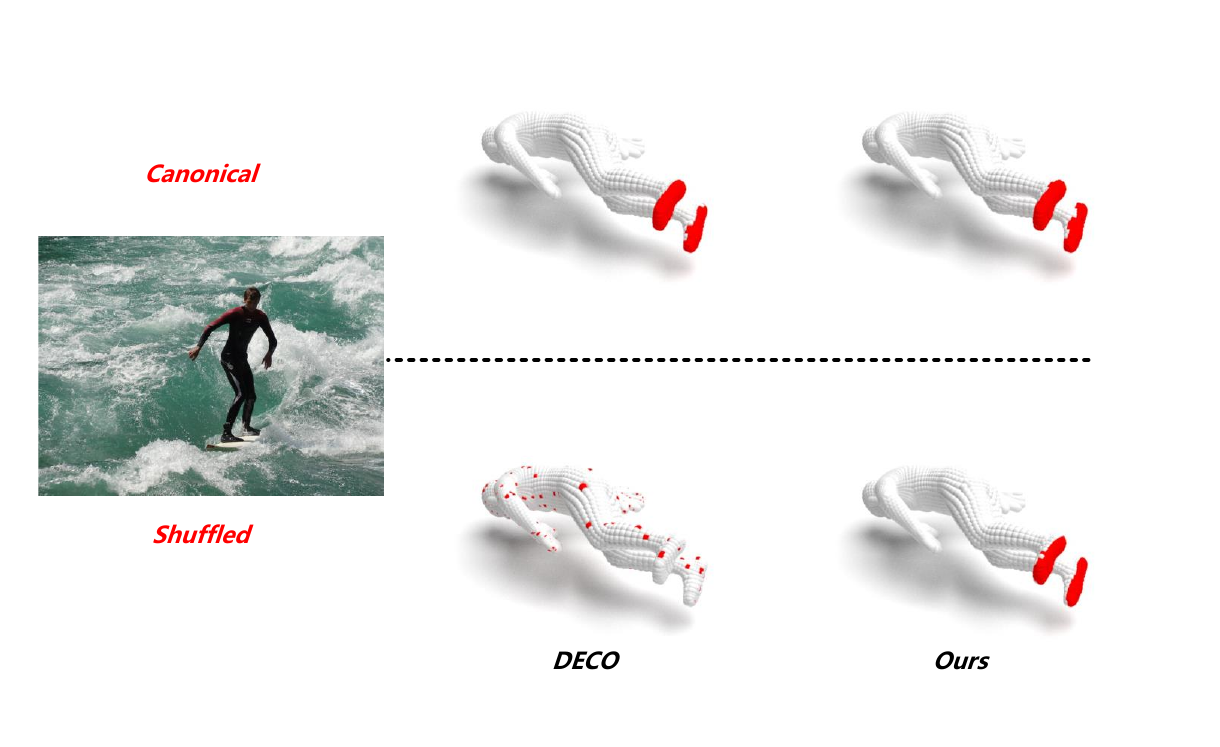}

        \end{overpic}
  \caption{Shuffling the sequence of smpl point cloud. The top row shows the contact prediction results on the original SMPL sequence point clouds, while the bottom row  shows the results after randomly shuffling the point cloud sequence. \textcolor[rgb]{0.9, 0.2, 0.2}{Red} indicates the points predicted as being in contact.}
  \label{fig:shuffle}
\end{figure}

\subsection{Extra experiment}
\label{sec 4.4:exp.Extra experiment}
\noindent\textbf{HPS with GRACE-Predicted Contacts.\ }Following the protocols established by DECO \cite{tripathi2023deco}  and HOT \cite{chen2023hot}, we investigate whether contact predictions from GRACE can enhance human pose and shape (HPS) regression. Experiments are conducted on the PROX quantitative dataset \cite{PROX:2019} under the same evaluation setup as DECO and HOT. PROX employs an optimization-based approach to fit SMPL-X bodies to images, assuming a known 3D scene and leveraging manually-annotated contact regions to encourage body-scene contact when sufficiently close, while penalizing interpenetration.To assess the effectiveness of GRACE, we replace the manually-annotated contact vertices with GRACE-predicted dense contact regions. For the "No contact" baseline, all contact constraints are disabled during optimization. The results, summarized in Tab. \ref{tab:exp_HPS}, demonstrate the impact of using GRACE-inferred contacts for HPS regression.

% Following DECO \cite{tripathi2023deco} and HOT \cite{chen2023hot}, we investigate whether contact predictions from GRACE benefit human pose and shape (HPS) regression. The evaluation is conducted on the PROX quantitative dataset \cite{PROX:2019} under the same experimental protocol as DECO and HOT. Results are summarized in Tab. \ref{tab:exp_HPS}.

\noindent\textbf{Shuffling the vertex order of the SMPL model point cloud.\ } Since point cloud data inherently lacks an explicit topological structure, shuffling the vertex order of an SMPL model does not alter its geometric shape. However, existing methods typically rely on learning a mapping between images and the inherent vertex sequences within structured human point clouds. Consequently, when the vertex order of the input point cloud is randomized, the contact prediction performance of these methods degrades significantly, rendering them unable to accurately identify dense contact regions (see Fig. \ref{fig:shuffle} DECO). In contrast, our approach establishes a direct mapping from geometric features to dense vertex contact probabilities, demonstrating inherent robustness to variations in point cloud vertex order. As long as the geometric structure of the point cloud remains unchanged, the contact prediction results remain stable and accurate, even when the vertex order is arbitrarily permuted (see Fig. \ref{fig:shuffle} Ours). This characteristic not only verifies the generalization capability of our model under unordered input conditions but also provides both theoretical and experimental support for its inference ability on unstructured or weakly structured human representations.

\noindent\textbf{Non-SMPL Series Human Mesh Contact Inference.\ }To validate the effectiveness of our method in handling unstructured human point clouds, we employ \textbf{Hunyuan 3D} \cite{hunyuan3d22025tencent} to infer detailed 3D point clouds from monocular images. Unlike structured human models such as SMPL and SMPL-X, the point clouds generated by Hunyuan 3D lack a predefined topological structure and exhibit variations in density as well as spatial distribution. This characteristic makes them a suitable benchmark for thoroughly evaluating the generalization capability of our framework.  After generating these diverse point clouds, we input them along with the corresponding images into the GRACE network, which has been trained on SMPL human point clouds from the DAMON dataset, to predict dense 3D contact regions. In Fig. \ref{fig:hunyuan}, we present the contact prediction results on unstructured human point clouds. As shown in the figure, even when the point cloud structure undergoes significant changes, GRACE maintains reasonable contact reasoning to a considerable extent, demonstrating its strong adaptability to diverse human representations.  

% \begin{table}[t]
% \footnotesize
% \centering
%   \renewcommand{\arraystretch}{1.}
%   \renewcommand{\tabcolsep}{6.pt}
%    \caption{HPS estimation performance using contact derived from different sources.}
%     \label{tab:exp_HPS}
% \begin{tabular}{c|cccccc}
% \toprule
%   \textbf{Methods}  & \ding{55} \textbf{Contact} & \textbf{Prox} & \textbf{HOT} & \textbf{DECO} & \textbf{GRACE} & \textbf{GT}\\ \midrule
% \textbf{V2V $\downarrow$} & 183.3 & 174.0 & 172.3 & 171.6 & \textbf{171.1} & 163.0 \\
% \bottomrule
% \end{tabular}
% \end{table}
% \begin{figure}[t]
%   \centering
%   \small
%         \begin{overpic}[width=0.92\linewidth]{figs/hunyuan_infer.pdf}

%         \end{overpic}
%   \caption{Contact estimation on non-smpl mesh.}
%   \label{fig:hunyuan}
% \end{figure}

Unlike existing methods, which suffer from inconsistent contact predictions due to the absence of a predefined topology, our framework preserves spatial coherence, making it applicable to various real-world scenarios—especially in cases where structured human models are unavailable.

\section{Conclusion}

We propose a novel paradigm for 3D dense human contact estimation, where contact estimation is defined as an implicit mapping between geometric features and contact probabilities, rather than a mesh topology-based sequential prediction task. This paradigm enables contact estimation to generalize to non-parametric human models, overcoming the limitation of previous methods that could only infer contacts for structured parametric models.To implement this paradigm, we introduce GRACE, a novel framework that effectively combines 2D interaction semantics with 3D geometric priors through a hierarchical cross-modal fusion mechanism. Furthermore, we enhance geometric error metrics to provide more comprehensive evaluation of contact prediction quality, and validate GRACE's superior performance across multiple benchmark datasets. We believe this study offers new insights into 3D dense human contact estimation and opens new research directions in this field.

\noindent\textbf{Limitations and Future Work.} Despite GRACE's strong performance, current training is limited to SMPL-family datasets due to data scarcity, resulting in reduced inference capability for point clouds significantly deviating from parametric human models (e.g., those with missing limbs). Future work will focus on constructing diversified human model-contact paired datasets to enhance GRACE's generalization. Another promising direction involves training multimodal large models for contact estimation tasks, aiming to develop more accurate and generalizable universal contact reasoning models.

\bibliographystyle{ACM-Reference-Format}
\bibliography{main}

\end{document}